\documentclass[10pt,twocolumn,letterpaper]{article}

\usepackage{cvpr} 
\usepackage{times}
\usepackage{epsfig}
\usepackage{booktabs} 
\usepackage{graphicx}
\usepackage{amsmath}
\usepackage{amssymb}
\usepackage{xcolor}
\usepackage{subcaption} 
\usepackage{comment}
\usepackage{makecell}

\usepackage[pagebackref,breaklinks,colorlinks,allcolors=blue]{hyperref}

\begin{document}

\title{Progressive Learning of a Diffusion-based Inpainting Model \\ for Separating Overlapped Fingerprints}


\author{
 Noor Hussein \qquad Anil K. Jain \qquad Karthik Nandakumar \\
 Michigan State University, East Lansing, MI 48824, USA \\
 {\tt\small \{hussei52, jain, nandakum\}@msu.edu}
}

\maketitle
\thispagestyle{empty}

\begin{abstract}
Overlapped friction ridge patterns are a recurring problem in latent fingerprints recovered from crime scenes and in live-scan scenarios where residual fingerprints on the sensor may corrupt subsequent acquisitions. Existing approaches for separating overlapped fingerprints either rely on rule-based orientation field completion that requires strong domain knowledge or train end-to-end deep neural networks that do not account for domain-specific considerations. This work introduces a diffusion-based pipeline for separating component fingerprints from an image containing overlapping friction ridge patterns. We formulate the separation problem as an inpainting task and progressively learn a diffusion model for this task in multiple stages. Starting from a pre-trained Stable Diffusion model, we progressively incorporate a fingerprint prior, add the ability to complete partial fingerprints, and finally propose \textbf{overlap-aware inpainting} that reconstructs each component print using a diffusion inpainting model based on multi-channel conditioning. Experiments on two public datasets demonstrate that component fingerprints reconstructed using the proposed diffusion-based inpainting method can match with their mated counterparts with very high probability.

\end{abstract}

 
\section{Introduction}
\label{sec:introduction}
Fingerprint recognition is one of the most widely used methods for recognizing individuals in diverse applications such as forensic identification, border control, mobile authentication, and large-scale civil registry systems. State-of-the-art fingerprint matchers can achieve high accuracy, even in the presence of noise, partial occlusion, distortion, low quality or low contrast~\cite{maltoni2009handbook, engelsma2019learning, grosz2023afr}. However, almost all matchers implicitly assume that the input fingerprint image has friction ridge patterns captured from a \emph{single} finger. 


One scenario where this breaks down is the case of overlapped fingerprints~\cite{chenjain2011separating}, where the region of interest contains two or more ridge patterns. While overlapped fingerprints are somewhat rare, they do occur occasionally in two operational settings. In forensic casework, latent fingerprints lifted from surfaces could be contaminated by an earlier impression left by a different finger or the same finger, resulting in a composite that an examiner must separate manually or discard~\cite{jain2010latent}. In live-scan acquisition, residual moisture or oil from a previous finger can leave a faint impression on the surface that mixes with the next finger captured, producing an overlap at the sensor itself. In both settings, the downstream matcher receives an image whose ridges are a function of two identities, not one.

\begin{figure}[t]
    \centering
    \includegraphics[width=0.9\columnwidth]{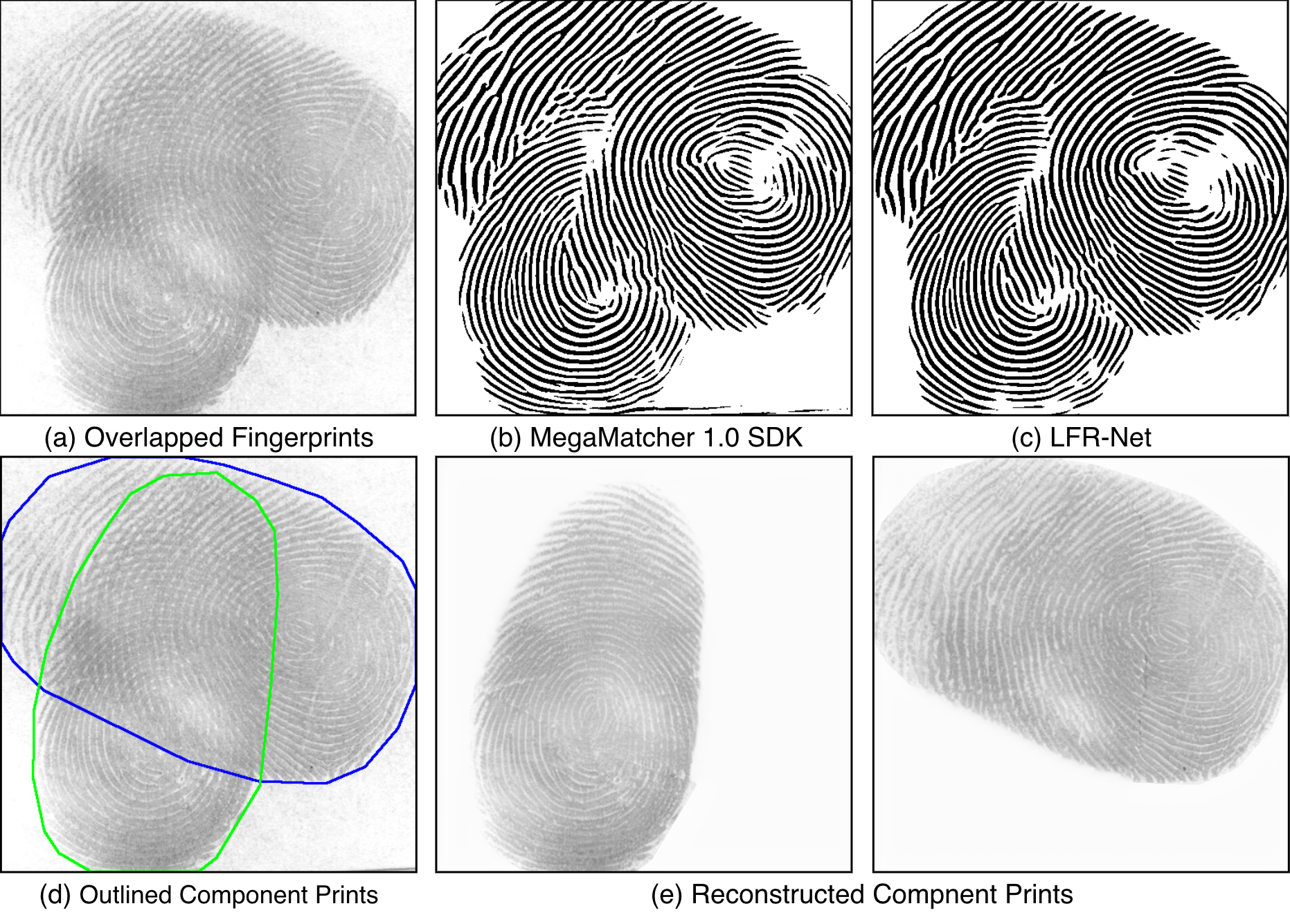}
    \caption{Existing fingerprint recognition systems are not capable of dealing with overlapped fingerprints. (a) Overlapped fingerprint image; (b) the ridge pattern extracted by MegaMatcher 1.0 SDK, (c) by LFR-Net~\cite{grosz2023latent} matchers; (d) overlapped image with component masks outlined; and (e) reconstructions using our method.}
    \label{fig:binary_images}
\vspace{-0.3cm}
\end{figure}

Since fingerprint matchers (e.g., MegaMatcher\footnote{\label{megamatcher}https://www.neurotechnology.com/megamatcher.html} and LFR-Net~\cite{grosz2023latent}) are explicitly designed to be robust against damaged, noisy, and partially occluded inputs, they do not necessarily fail when  presented with an overlapped fingerprint image. Instead, they confidently extract information that mixes minutiae and ridges of both contributing fingers (see Figure~\ref{fig:binary_images}). The resulting ridge structure is often considered to be a valid, continuous single ridge flow pattern, despite failing to correspond to any single true identity. During recognition, this composite probe can trigger false matches against either contributing identity, or even a third unrelated identity whose minutiae happen to align with the mixed pattern. This increases the False Match Rate (FMR), a liability in forensic deployments where wrongful identification may have severe consequences. In essence, robustness of the matcher may turn into a limitation when the input image violates the single-identity assumption.

A fingerprint recognition system capable of dealing with overlapped fingerprints should perform three tasks, namely, \textit{detection}, \textit{localization}, and \textit{separation}. Given a generic fingerprint image, the detector's goal is to determine whether the input image contains overlapped fingerprint patterns. Once an overlapped fingerprint is detected, the objective of the localization module is to determine the regions containing only the component fingerprints, as well as identify the overlapping region. Finally, the separation module aims to reconstruct the component fingerprints in the overlapping region. In this work, \textit{we focus solely on the separation step}.

Existing methods for separating overlapped fingerprints fall under two main paradigms. Model-based methods~\cite{zhaojain2012modelbased} formulate separation as an orientation field completion task using relaxation labeling and hand-crafted priors over ridge geometry. Their performance degrades in high-overlap, low-clarity regions where separation is most needed. End-to-end learning-based methods such as FinSNet~\cite{yoo2020finsnet} train a deep neural network to map an overlapped fingerprint image to a single target fingerprint component. While this approach sidesteps explicit orientation estimation, it has other limitations: training data is entirely based on synthetic fingerprint images and the model has no explicit understanding of what a plausible friction ridge pattern looks like. 


Recent synthetic fingerprint image generation models (e.g., latent diffusion models (LDMs) such as GenPrint~\cite{grosz2024universal}) have the potential to address these limitations because they already have a strong fingerprint prior. A pretrained, high-fidelity ridge-texture prior constrains the target solution space to the manifold of valid single-finger impressions, effectively regularizing the under-determined inverse problem of overlapped fingerprint separation. Based on this insight, we pose overlapped fingerprint separation as a conditional inpainting task in the latent diffusion space. This work makes the following contributions:

\begin{itemize}
    \item We formulate overlapped fingerprint separation as a multi-channel conditional diffusion inpainting task in the latent space, leveraging a stacked low-rank adapter (LoRA) architecture to update the diffusion model.
    \item We introduce an auxiliary fine-tuning mechanism including a joint composition loss to couple the paired denoising chains, enforcing mutual consistency between the reconstructed component fingerprints.
    \item We construct a large-scale, paired training data of 55,000 highly diverse synthetic overlapped fingerprints built from real single finger impressions across 11 fingerprint databases, keeping the component prints real and simulating only the overlap process.

\end{itemize}

\section{Related Work}

Separation of overlapped fingerprints has traditionally been approached as a ridge orientation decomposition problem, in which the mixed ridge flow of an overlapped fingerprint image is separated into two underlying ridge patterns. 

\textbf{Classical methods.} Chen et al.~\cite{chenjain2011separating} used local Fourier analysis and relaxation labeling to iteratively assign ridge ownership within the overlap region. Their work explored both a core-point-guided formulation and a formulation that avoided explicit singular point assumptions. Shi et al.~\cite{shi2011separating} further investigated ridge separation using orientation consistency and structural ridge constraints. Zhao and Jain~\cite{zhaojain2012modelbased} further formulated the problem as orientation field reconstruction, estimating the dominant ridge flow of each component fingerprint before reconstructing ridge textures through model-based enhancement. Zhang et al.~\cite{zhang2014overlapped} proposed adaptive orientation field fitting to better handle severe overlap conditions.
Although these approaches are interpretable and exploit important domain knowledge about ridge geometry, they depend strongly on accurate orientation estimation and handcrafted reconstruction heuristics. Their performance degrades substantially in regions of high overlap, poor ridge clarity, or inconsistent local ridge flow, where reliable orientation estimation becomes difficult.

\textbf{Learning-based separation methods.} Machine learning models can also be trained to map overlapped prints directly to clean component print images, bypassing explicit orientation field estimation. Yoo et al.~\cite{yoo2020finsnet} introduced FinSNet, a U-Net trained end-to-end to isolate targeted fingerprint from an overlapping composite using spatial transformations and region of interest marking. Although these learned methods exhibit greater inference speed and robustness to noise than classical estimators, they are trained on simplified synthetic texture patterns (e.g. SFinGE~\cite{cappelli2004sfinge}). Consequently, these networks lack a strong structural prior and often produce unrealistic ridge patterns.

\textbf{Diffusion inpainting and fingerprint generation.} Denoising Diffusion Probabilistic Models (DDPMs)~\cite{ho2020denoising} have significantly advanced the state-of-the-art in conditional image synthesis and editing. GenPrint~\cite{grosz2024universal} introduced a controllable LDM for fingerprint generation, effectively learning a highly realistic structural prior over friction ridge manifolds. In image guided editing, architectures such as Paint-by-Example~\cite{yang2023paint} extend latent diffusion by injecting extra U-Net input channels to ingest bounding masks and conditional reference images, enabling localized context-aware inpainting. This work aims to exploit these advancements and repurpose diffusion-based exemplar inpainting to disentangle the intertwined ridge topologies already present within an overlapped region, rather than synthesizing new ridges out of context.

\section{Methodology}

The goal of this work is to train a generative model for separating overlapped fingerprints. We assume that an overlapped fingerprint has already been detected and that the overlapping region has been localized. 

\noindent \textbf{Notations}: Let $I \in [0,1]^{H \times W}$ denote a grayscale fingerprint image, where $0$ and $1$ indicate black and white pixels, respectively. Let $M \in \{0,1\}^{H \times W}$ be the corresponding binary mask, where $0$ indicates the background region and $1$ indicates the foreground fingerprint region. In general, we assume that pixels in the background region of a fingerprint image are white, i.e., $I(p) = 1$ when $M(p) = 0$, where pixel $p := (h,w), ~ h \in [1,H]$ and $w \in [1,W]$. If a given fingerprint image does not satisfy this assumption, we can apply a segmentation algorithm to enforce it.


\subsection{Problem Formulation}

Let $I_O$ be an overlapped fingerprint image formed by composition of two underlying fingerprints $I_A$ and $I_B$ (see Figure \ref{fig:overlap_example_outline}). Let $M_A$ and $M_B$ be the binary foreground masks corresponding to the component prints $I_A$ and $I_B$, respectively. Thus, the overlapped fingerprint can be modeled as: 

\begin{figure}[t]
    \centering
    \includegraphics[width=0.8\columnwidth]{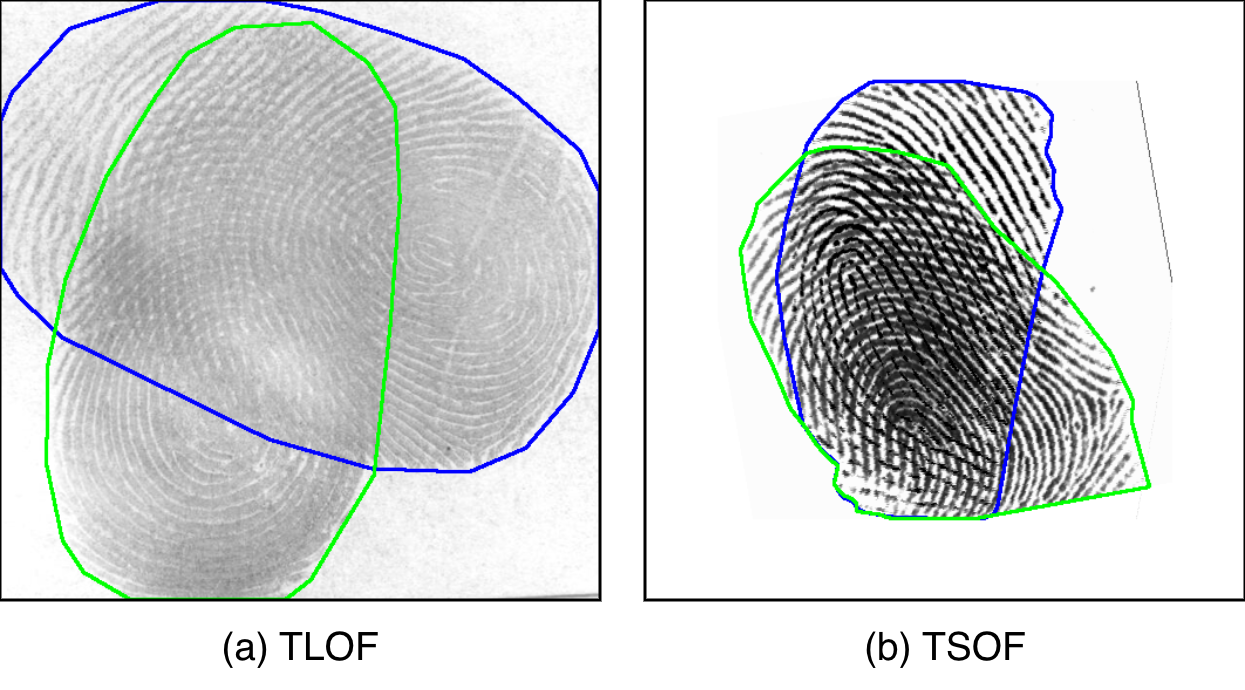}
    \caption{Examples of overlapped fingerprints from two datasets: (a) Tsinghua Latent Overlapped Fingerprint (TLOF) and (b) Tsinghua Simulated Overlapped Fingerprint (TSOF). The \textcolor{green}{green} and \textcolor{blue}{blue} outlines show the masks \textcolor{green}{$M_A$} and \textcolor{blue}{$M_B$} of the component fingerprints $I_A$ and $I_B$, respectively. The intersection between the two masks \textcolor{green}{$M_A$} and \textcolor{blue}{$M_B$} represents the overlapping region $M_O$.}
    \label{fig:overlap_example_outline}
\vspace{-0.2cm}
\end{figure}

\begin{figure*}[h]
    \centering
    \includegraphics[width=\linewidth]{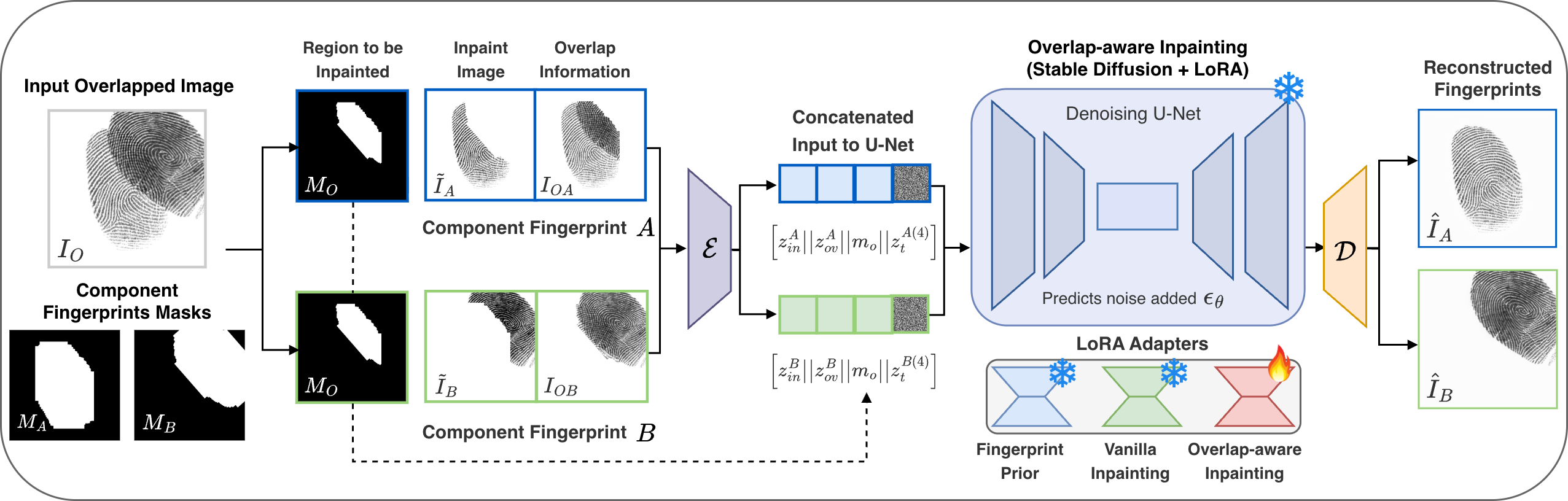}
    \caption{Illustration of the proposed overlap-aware inpainting method (Stage 2).}
    \label{fig:pipeline}
    \vspace{-0.2cm}
\end{figure*}

\begin{equation}
    I_O(p) = \begin{cases}
        g(I_A(p),I_B(p)) & M_A(p) = 1 ~\text{and}~ M_B(p) = 1\\
        I_A(p)           & M_A(p) = 1 ~\text{and}~ M_B(p) = 0 \\
        I_B(p)           & M_A(p) = 0 ~\text{and}~ M_B(p) = 1 \\
        1                & M_A(p) = 0 ~\text{and}~ M_B(p) = 0,
    \end{cases}
\end{equation}

\noindent where $g$ is the (typically unknown) mixing function that entangles the component fingerprints in the overlapping region. Given $I_O$ together with the ground-truth binary masks $M_A$ and $M_B$ (obtained by manually segmenting each fingerprint instance), our goal is to recover $\hat{I}_A \approx I_A$ and $\hat{I}_B \approx I_B$ that do not contain mixed ridges and are close approximations of the component fingerprints. 

Let $M_O = M_A \& M_B$ be the mask that indicates the overlapping region in $I_O$, where $\&$ denotes the logical-AND operation. The separation challenge is concentrated in $M_O$, where the ridges of both fingerprints are mixed and the algorithm must \emph{separate} or \emph{unmix} them. Typically, the difficulty depends on the amount of overlap and the nature of the mixing function. When the overlapping region is large, the problem becomes more difficult because large portions of the component fingerprints need to be reconstructed. The difficulty is also high when the mixing function obfuscates both component fingerprints in the overlapping region.    

We formulate the separation/unmixing problem as a conditional inpainting task using diffusion models~\cite{yang2023paint}. The overlap region $M_O$ is treated as a structured ``hole'' whose contents must be filled with ridges that (i) lie on the manifold of plausible single-finger impressions under a generative fingerprint prior, (ii) extend the ridges visible in the non-overlapping region of the component fingerprints, and (iii) are jointly consistent with the observed $I_O$ when re-composed based on the individual reconstructions. 

\subsection{Unmixing via Fingerprint Inpainting}

Our overlapped fingerprint separation algorithm is based on a latent diffusion model constructed starting from the Stable Diffusion v1.5~\cite{rombach2022high} (SD1.5) model. The diffusion model consists of the following components: an image encoder $\mathcal{E}$ that maps $512 \times 512$ images to a $64 \times 64 \times 4$ latent space, a CLIP text encoder, a U-Net module $\epsilon_\theta$ that performs denoising in the latent space, and an image decoder $\mathcal{D}$ that maps a latent back to the image space. In this work, the text prompt is always left empty. Hence, the network operates as an unconditional latent diffusion model from the language perspective and is conditioned purely through additional input channels in the latent space. The proposed unmixing model is assembled from the base SD1.5 model in three stages of specialization, with each stage adding a low-rank adapter (LoRA) ~\cite{hu2022lora} to the attention projections of every cross- and self-attention block in the U-Net. 

\noindent \textbf{Stage 0: Incorporating fingerprint domain prior}. The goal of this stage is to equip the SD1.5 model with the knowledge of what a fingerprint looks like, before teaching it to complete missing fingerprint regions. This is achieved by leveraging the parameters of an existing synthetic fingerprint generation model called \emph{GenPrint}~\cite{grosz2024universal}. The GenPrint model was trained on a large corpus of single-finger impressions and it uses rank-4 LoRA adapters in conjunction with a base SD1.5 model for the purpose of synthetic fingerprint generation. We extract these LoRA parameters from GenPrint and \emph{fuse} them into the base SD1.5 weights to incorporate the fingerprint domain prior. These modified weights are frozen for all subsequent stages. 

\noindent \textbf{Stage 1: Vanilla inpainting for single fingerprint completion}. The objective of this stage is to train the model to reconstruct a single fingerprint image from its partial observation. Let $I_S$ be a single fingerprint image with corresponding foreground mask $M_S$. We synthesize a hole $M_H$ (soft ellipse) within the foreground region of $I_S$ ($M_H$ is contained within $M_S$) and set the pixels of $I_S$ inside the hole to value $1$. Let $\widetilde{I}_S$ denote the single fingerprint image with the synthesized hole. Given $\widetilde{I}_S$ and $M_H$, we now train the diffusion model to recover $\hat{I}_S \approx I_S$. 

To enable the diffusion model to learn this vanilla inpainting task, the U-Net input convolution is expanded from 4 to 9 channels by concatenating the latent encoding of the partial fingerprint (4 channels) and the down-sampled hole mask (1 channel) to the standard context latent (4 channels). The weights corresponding to the 5 new channels are initialized with 0, so that the expanded U-Net is functionally identical to the base U-Net at initialization. The full input to the denoiser at timestep $t$ is:

\begin{equation} 
    z_t^{(9)} \;=\; \big[z_{in} \;\big|\big|\; m_h  \;\big|\big|\;  z_t^{(4)} \,\big],
\end{equation}
where $z_{in} = \mathcal{E}(\widetilde{I}_S) \in \mathbb{R}^{4 \times 64 \times 64}$ is the latent image encoding of the partial fingerprint image $\widetilde{I}_S$, $m_h \in \mathbb{R}^{1 \times 64 \times 64}$ is the bi-linearly down-sampled version of binary hole mask $M_H$, $z_t^{(4)} \in \mathbb{R}^{4 \times 64 \times 64}$ is the standard 4-channel noisy (context) latent of the target fingerprint image to be reconstructed, and $\big|\big|$ denotes channel-wise concatenation. New rank-16 LoRA blocks are added to the attention layers and trained using the standard DDPM~\cite{ho2020denoising} noise-prediction (mean squared error) loss. After training, both the LoRA and the 9-channel convolution parameters are incorporated into the base model and carried forward to Stage 2.


\noindent \textbf{Stage 2: Overlap-aware inpainting for overlapped fingerprint separation}. In this stage, the model is trained to reconstruct two component fingerprints from the overlapped fingerprint. Let $\widetilde{I}_A$ ($\widetilde{I}_B$) be the non-overlapped ridge pattern of the component fingerprint $I_A$ ($I_B$) obtained from the given overlapped image $I_O$ by applying the foreground mask $M_A\&\bar{M}_O$ ($M_B\&\bar{M}_O$), where $\bar{M}_O = (1-M_O)$. Let $I_{OA}$ ($I_{OB}$) be the partially overlapped ridge pattern of the component fingerprint $I_A$ ($I_B$) obtained from the overlapped image $I_O$ by applying the foreground mask $M_A$ ($M_B$). Using these non-overlapped and partially overlapped versions of the component fingerprints, we attempt to reconstruct the component fingerprints.   

To enable the diffusion model to learn this overlap-aware inpainting task, the U-Net input convolution is further expanded from 9 to 13 channels by concatenating the latent encoding of the non-overlapped component fingerprint (4 channels), latent encoding of the partially overlapped component fingerprint (4 channels), and the down-sampled overlap mask (1 channel) to the standard context latent (4 channels). The weights corresponding to the 4 new channels are initialized with 0, so that the expanded U-Net is functionally identical to the Stage 1 U-Net at initialization. The full input to the denoiser at timestep $t$ is:

\begin{equation} 
    z_t^{(13)} \;=\; \big[z^{*}_{in} \;\big|\big|\; z^{*}_{ov} \;\big|\big|\; m_o  \;\big|\big|\;  z_t^{*(4)} \,\big],
    \label{eq:latent_tensors}
\end{equation}

\noindent where $* \in \{A,B\}$, $z_{in}^{*} = \mathcal{E}(\widetilde{I}_*) \in \mathbb{R}^{4 \times 64 \times 64}$ is the latent image encoding of the non-overlapped component fingerprint $\widetilde{I}_*$, $z_{ov}^{*} = \mathcal{E}(I_{O*}) \in \mathbb{R}^{4 \times 64 \times 64}$ is the latent image encoding of the partially overlapped component fingerprint $I_{O*}$, $m_o \in \mathbb{R}^{1 \times 64 \times 64}$ is the bi-linearly down-sampled version of binary overlap mask $M_O$, and $z_t^{*(4)} \in \mathbb{R}^{4 \times 64 \times 64}$ is the standard 4-channel noisy (context) latent of the target fingerprint image $I_*$ to be reconstructed. Visualization of conditioning inputs are shown in Figure \ref{fig:pipeline} along with an illustration of the proposed pipeline. 

The above conditioning approach can be read as a fingerprint specific adaptation of Paint-by-Example~\cite{yang2023paint}, rather than supplying a separate exemplar image through a dedicated encoder branch, we exploit the fact that the ``exemplar'' for each component fingerprint is already partially present in the overlapped fingerprint. The model is given the clean part of the target via $z_{in}$, the location of the corrupted region via $m_o$, and the mixed ridges inside that region via $z_{ov}$. At each pixel in the overlapping region, the model must decide which ridges belong to the target finger.

The overlap-aware inpainting (Stage 2) stacks a new set of rank-16 LoRA blocks on top of the merged Stage 0 and 1 weights. Since each overlapped image admits two valid reconstruction targets (corresponding to the two component fingerprints), we employ a batched-paired forward. In other words, the conditioning tensors for both component fingerprints are stacked along the batch dimension, so that a single U-Net forward produces both reconstructions simultaneously. Beyond halving the U-Net evaluations per step, this layout also makes it inexpensive to add coupling losses between the two reconstructions (e.g., re-composing them and comparing to $I_{O}$). We train Stage 2 just like Stage 1 with the standard DDPM noise-prediction objective.
\begin{equation}
    \mathcal{L}_{\text{diff}} = 
    \mathbb{E}_{t,\, \epsilon,\, (I_O, I_A, I_B)}
    \Big[\, \big\| \epsilon - \epsilon_\theta\!\left(z_t^{(13)}, t\right) \big\|_2^2 \,\Big],
\end{equation}
computed simultaneously over both component reconstructions through the batched-paired forward. The model is trained for 48,000 steps on 4 NVIDIA RTX 6000 Ada GPUs (batch size 4/GPU) with AdamW (learning rate $2 \times 10^{-4}$, cosine schedule, 500 warm-up steps, weight decay 0.01, gradient clip 1.0) and \texttt{bf16} precision.

\noindent \textbf{Fine-tuning with auxiliary losses}: The diffusion objective $\mathcal{L}_{\text{diff}}$ supervises the model in latent space at every timestep. However, it provides only indirect guidance about the final pixel space reconstruction and does not enforce that the two component output fingerprints are jointly consistent with the observed overlap image $I_O$. Therefore, we fine-tune the trained Stage 2 checkpoint with two additional losses that act directly on the decoded reconstructions, while retaining $\mathcal{L}_{\text{diff}}$.
Both additional losses operate on a one-step DDIM denoising of the clean target. At a randomly sampled timestep $ t \leq 200 $ in the low noise regime (where one-step estimate is reliable), the clean latent is estimated as:
\begin{equation}
    \hat{z}_0^* = \frac{z^*_t - \sqrt{1-\bar\alpha_t} \, \hat{\epsilon}_\theta}{\sqrt{\bar\alpha_t}},
\end{equation}
where $\bar\alpha_t$ is the cumulative noise schedule coefficient~\cite{ho2020denoising} at timestep $t$ and $\hat{\epsilon}_{\theta}$ is the predicted noise. The estimated latent is decoded using the image decoder to obtain $\hat{I}_{*}  = \mathcal{D}(\hat{z}^*_0)$. 
\begin{figure}[t]
    \centering
    \includegraphics[width=\columnwidth]{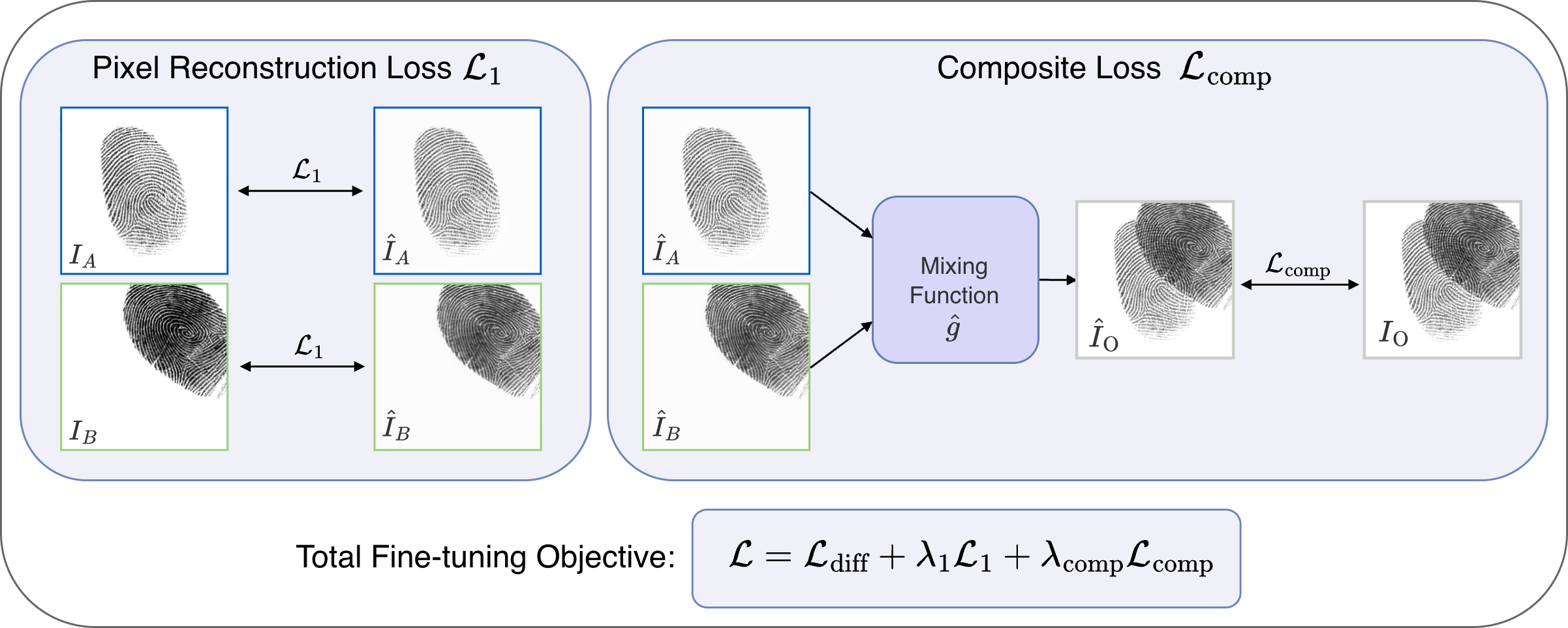}
    \caption{Auxiliary losses for fine-tuning of Stage 2 model.}
    \label{fig:finetuning}
    \vspace{-0.2cm}
\end{figure}

We supervise the decoded reconstruction in the pixel space with $\ell_1$ loss over the full fingerprints: 
\begin{equation}
    \mathcal{L}_1 = \tfrac{1}{2}\sum_{* \in \{A, B\}} \frac{1}{s(M_*)} \sum_{M_*(p)=1} \big|\hat{I}_{*}(p) - I_*(p)\big|,
\end{equation}
where $s(.)$ represents the mask size/area (number of pixels with value $1$). Note that the loss is computed over the entire mask of the fingerprint $M_*$ rather than only the overlap region $M_O$. $\mathcal{L}_1$ is applied to both component fingerprint reconstructions and averaged. 

The two reconstructions $\hat{I}_{A}$ and $\hat{I}_{B}$ are not independent and must jointly be consistent with the given overlapped image $I_{O}$. Using the batched-paired forward, we re-compose the decoded images into a reconstruction of the overlapped fingerprint image and penalize the discrepancy with the input overlapped image.

\begin{equation}
    \mathcal{L}_{\text{comp}} = \big\| \hat{I}_{O} - I_{O} \big\|_1,
\end{equation}
where 
\begin{equation}
    \hat{I}_O(p) = \begin{cases}
        \hat{g}(\hat{I}_A(p),\hat{I}_B(p)) & M_A(p) = 1 ~\text{and}~ M_B(p) = 1\\
        \hat{I}_A(p)           & M_A(p) = 1 ~\text{and}~ M_B(p) = 0 \\
        \hat{I}_B(p)           & M_A(p) = 0 ~\text{and}~ M_B(p) = 1 \\
        1                & M_A(p) = 0 ~\text{and}~ M_B(p) = 0,
    \end{cases}
\end{equation}
where $\hat{g}$ is a surrogate mixing function (described later in Section \ref{sec:TrainingDataPrep}). This term couples the two parallel denoising paths into a single joint optimization. Thus, the total fine-tuning loss when $t\leq200$ is:
\begin{equation}
    \mathcal{L} = \mathcal{L}_{\text{diff}} + \lambda_1\mathcal{L}_1 + \lambda_{\text{comp}} \mathcal{L}_{\text{comp}},
\end{equation}
with $\lambda_1 = 1.0$ and $\lambda_{\text{comp}} = 0.5$. For timesteps $t > 200$, the training is supervised only by $\mathcal{L}_{\text{diff}}$. 


\subsection{Inference}

During inference, we draw $z^*_T \sim \mathcal{N}(0, I)$ and run $T=50$ DDIM~\cite{song2021ddim} denoising steps. The conditioning tensors $z^*_{in}$, $m_o$, and $z^*_{ov}$ are prepared based on $I_O$, $M_A$ and $M_B$ as shown in Eq. \ref{eq:latent_tensors}. We run one denoising chain per component, yielding $\hat{I}_{A}$ and $\hat{I}_{B}$, which are then passed to a matcher or a human fingerprint examiner for evaluation.

\subsection{Training Data Preparation}
\label{sec:TrainingDataPrep}
For both training stages, we synthesize the required training set from real single-finger slap/plain impressions. We aggregate source fingerprints from 11 databases: FVC2000 DB2~\cite{maio2002fvc2000}, FVC2002 DB1~\cite{maio2002fvc2002} (excluding identities 101-110)\footnote{Since the test set TSOF was created using impressions from identities 101-110 in FVC2002-DB1, these impressions were excluded from our training data preparation.}, FVC2004 DB1~\cite{maio2004fvc2004}, IIITD MOLF~\cite{sankaran2015molf}, IIITD MUST~\cite{malhotra2023must}, IIITD SLF~\cite{sankaran2012iiitdslf}, Jaipur MNIT~\cite{deb2018jaipurmnit}, LFIW~\cite{liu2024lfiw}, NIST SD300~\cite{fiumara2018nist300}, NIST SD301~\cite{fiumara2018nist301}, and NIST SD302~\cite{fiumara2019nist302}. Several of these databases are released in multiple subsets (e.g., different sensors or capture conditions), so we treat each subset independently and sample from 22 subsets in total. All images are converted to grayscale and cropped or padded to $512 \times 512$ to retain their original 500ppi resolution. A pre-trained SqueezeUNet from~\cite{grosz2023afr} was used to produce a binary mask $M$ for each fingerprint. 

\noindent \textbf{Stage 1 Training}: Vanilla inpainting is trained on all 50,014 single impressions from the aggregated databases. The holes are synthesized as ellipses, and their area, aspect ratio, position, and orientation are randomized. To make the vanilla inpainting task progressively harder, curriculum learning is employed and the average hole area is gradually increased from 15\% to 60\% as training progresses.

\noindent \textbf{Stage 2 Training}: The available public-domain overlapping fingerprint datasets such as TLOF \cite{feng2012robust} and TSOF \cite{chenjain2011separating} have limited sample sizes, which are not sufficient to train the proposed model. Hence, we hold out TLOF and TSOF as test sets and synthesize our own training data from the aggregated single-fingerprint databases. To prevent any single subset from dominating, we draw 50 fingerprints from each of the 22 subsets and form 2,500 ordered pairs per subset.

For each ordered pair of fingerprint images $(I_A, I_B)$ corresponding to different identities and drawn randomly from each database, we synthesize an overlapping image as follows. We first sample the affine transform parameters $(\phi, t_x, t_y)$ such that the rotation angle $\phi \sim \mathcal{U}[0, 2\pi)$ and the translation parameters $t_x, t_y \sim \mathcal{U}[-120, 120]$ pixels, where $\mathcal{U}$ denotes the uniform distribution. We then apply the affine transformation (rotation about the image center followed by translation) to $I_B$ and $M_B$, filling the regions outside the masks with white pixels. For notational simplicity, we shall henceforth use $I_B$ and $M_B$ to denote the transformed versions of the second component fingerprint. We compute $M_O = M_A \& M_B$ and keep the pair only if
\begin{equation}
    r \;=\; \frac{s(M_O)}{\min(s(M_A),\, s(M_B))} \;\in\; [0.15,\, 1.0].
\end{equation}
Pairs not satisfying this constraint are re-sampled up to ten times and discarded otherwise. We compose the overlapped fingerprint $I_O$ using one of the following three mixing functions randomly selected for each pair with probabilities $\{0.50, 0.35, 0.15\}$, respectively: 
\begin{itemize}
    \item \emph{minimum pixel}: $g(x_1,x_2) = \min(x_1, x_2)$, mimicking ink deposition where the darker (more ink) impression dominates,
    \item \emph{alpha blending}: $g(x_1,x_2) = \alpha x_1 + (1-\alpha)x_2$ with $\alpha \sim \mathcal{U}[0.35, 0.65]$,
    \item \emph{weighted summation}: $g(x_1,x_2) = w \min(x_1,x_2) + (1-w)\mathrm{mean}(x_1,x_2)$ with $w \sim \mathcal{U}[0.5, 0.9]$, a softer variant of \emph{minimum pixel}.
\end{itemize}
Each accepted tuple $(I_O, I_A, I_B, M_A, M_B, M_O, \phi, t_x, t_y)$ is stored on disk and yields two training targets $I_A$ and $I_B$. We synthesize 55,000 unique pairs and partition them 95\%/5\% for training/validation. The overlap ratio $r$ spans [0.15, 1.0] with a mean of 0.68. 


\section{Experiments and Results}

\subsection{Experimental Setup}

\noindent \textbf{Test sets}. Following previous literature, we evaluate on two public datasets for overlapped fingerprints, both released by Tsinghua University. The Tsinghua Latent Overlapped Fingerprint Dataset (TLOF)~\cite{feng2012robust} consists of 100 real latent overlaps enhanced and lifted from white paper. The dataset is released with component masks, but we observed that several of these masks cover only a portion of the visible ridge area, we therefore re-annotated masks covering the full print of each component fingerprint. 
The identities come from 12 plain fingerprints that are used as templates. The Tsinghua Simulated Overlapped Fingerprint dataset (TSOF)~\cite{chenjain2011separating} consists of 100 synthetic overlaps from 10 identities (101–110) of FVC2002 DB1, obtained by overlapping impression \#3 with \#4 of all 10 fingers; impression \#1 serves as gallery templates.
Since the dataset was released without component masks, we manually annotated the masks. 
Example images are shown in Figure \ref{fig:overlap_example_outline}.

\noindent \textbf{Evaluation protocol.} For each overlapped fingerprint image, two component fingerprints (henceforth referred to as probes) are produced. These probes are matched against a gallery of clean plain fingerprint templates, with 10 and 12 fingerprints for the TSOF and TLOF datasets, respectively. Both verification (one-to-one matching) and closed-set identification are performed using \textbf{MegaMatcher 1.0 SDK}\textsuperscript{\ref{megamatcher}}. For verification, we report the True Match Rate (TMR) at a threshold of $48$, which corresponds to $0.01$\% FMR according to the MegaMatcher 1.0 SDK manual. For closed-set identification, we output all identities whose match score exceeds a threshold of 48 and report the True Positive Identification Rate (TPIR) (correct identity is included in the candidate set) and False Positive Identification Rate (FPIR) (incorrect identity is included in the candidate set). We also expand the gallery to 2,022 identities (2,000 from NIST SD4, 12 from TLOF and 10 from TSOF), run closed-set identification using the MegaMatcher and \textbf{LFR-Net}~\cite{grosz2023latent} matchers, and report the rank-1 accuracy.

\noindent \textbf{Benchmarking scenarios}: We consider five benchmarking scenarios: (i) \textit{No Unmixing}: In this scenario, no processing is done and the given overlapping fingerprint image $I_O$ is directly output as the reconstructed component fingerprints. (ii) \textit{Naive Unmixing}: The partially overlapped images $I_{OA}$ and $I_{OB}$ (obtained by applying the masks $M_A$ and $M_B$ to $I_O$) are output as the reconstructed component fingerprints $\hat{I}_A$ and $\hat{I}_B$. (iii) \textit{Vanilla inpainting}: The diffusion model checkpoint after Stage 1 training is used to separate overlapped fingerprints in the following way. We consider $\widetilde{I}_A$ and $\widetilde{I}_B$ as partial single fingerprint images, $M_O$ as the hole to be filled in each image, and obtain the reconstruction of the component fingerprints using the Stage 1 model. (iv) \textit{Overlap-aware inpainting}: This scenario uses the model after Stage 2 training (without any finetuning based on  auxiliary losses) for reconstruction. (v) \textit{Overlap-aware inpainting + finetuning}: The final model after Stage 2 training and finetuning is used for reconstruction.


\noindent \textbf{Baseline methods.} Since the matchers used in earlier works~\cite{chenjain2011separating, feng2012robust,yoo2020finsnet} are older versions of VeriFinger SDK that are no longer available, we cannot directly compare the proposed method against these baselines quantitatively. However, we qualitatively compare our results with the previous methods in the supplementary material.

\subsection{Closed-set Identification Results}

Table \ref{tab:tar_far} reports the closed-set identification performance on TSOF and TLOF based on MegaMatcher. Two noteworthy trends are observable. Firstly, although the
no unmixing scenario appears to achieve decent TPIR with FPIR of $0\%$ (since the raw probe contains ridges from \emph{both}
genuine identities $I_A$ and $I_B$, matching either counts as a hit), this apparent robustness disappears in the rank-1 setting
(Table~\ref{tab:rank1}), where matching drops to 52.6\% on TSOF and 56.6\% on TLOF. Naive unmixing partially mitigates this via the target's mask, but residual ridges from the other component leak through, resulting in FPIR (4.28\% and 1.50\%) unacceptably high. Specialized separation algorithms are required to reduce FPIR to acceptable levels.
Secondly, the relatively high TPIR of naive unmixing shows that existing matchers are robust enough to deal with partially corrupted single fingerprints, especially when the overlap ratio is small (as in TLOF dataset). However, relying solely on the generative models to reconstruct corrupted regions is also problematic because it can lead to large hallucinations even in the non-overlapping regions. This explains the low TPIR of vanilla inpainting, especially on TSOF where the overlap ratio is typically large. The proposed overlap-aware inpainting hits the right balance between preserving ridge information in the non-overlapping regions and reconstructing the overlapped regions in a consistent manner.  

\begin{table}[h!]
\centering
\caption{Closed-set identification results on TSOF and TLOF. Note that TPIR ($\uparrow$) and FPIR ($\downarrow$) \% are computed at a threshold of 48.} 
\vspace{0.1cm}
\label{tab:tar_far}
\small

\begin{tabular}{@{}l cc cc@{}}
\toprule
& \multicolumn{2}{c}{\textbf{TSOF}} & \multicolumn{2}{c}{\textbf{TLOF}} \\
\cmidrule(lr){2-3} \cmidrule(lr){4-5}
\textbf{Method} & TPIR & FPIR & TPIR & FPIR \\
\midrule
No unmixing          & 77.4 & 0.00 & 73.4 & 0.00 \\ 
Naive unmixing               & 78.0 & 4.28 & \textbf{95.5} & 1.50 \\
\midrule
Vanilla inpainting                & 15.0 & 0.00 & 65.0 & 0.00 \\
Overlap-aware inpainting        & 89.0 & 0.17 & 93.5 & 0.05 \\
\quad + fine-tuning & \textbf{90.5} & 0.17 & 95.0 & 0.00 \\
\bottomrule
\end{tabular}
\vspace{-0.2cm}
\end{table}

Table \ref{tab:rank1} reports rank-1 accuracy based on an expanded gallery containing 2,022 identities (2000 NIST SD4, 12 TLOF, 10 TSOF). The proposed method outperforms the naive unmixing strategy for both matchers by a large margin (especially on TSOF with large overlap). This shows that the separation algorithm is beneficial irrespective of the strengths of the fingerprint matcher deployed. Furthermore, the results are also compared against FinSNet~\cite{yoo2020finsnet}. It must be emphasized that this comparison may not be fair because FinSNet uses an older VeriFinger SDK. Hence, superior performance of the proposed method could be due to a combination of better matcher and  separation algorithm.   

\begin{table}[t]
\centering
\caption{Rank-1 accuracy (\%) results with an expanded gallery.}
\label{tab:rank1}
\small
\setlength{\tabcolsep}{4pt}
\renewcommand{\arraystretch}{1.15}
\begin{tabular}{@{}l cc cc@{}}
\toprule
& \multicolumn{2}{c}{\textbf{MegaMatcher}} & \multicolumn{2}{c}{\textbf{LFR-Net}} \\
\cmidrule(lr){2-3} \cmidrule(lr){4-5}
\textbf{Method} & \textbf{TSOF} & \textbf{TLOF} & \textbf{TSOF} & \textbf{TLOF} \\
\midrule
No unmixing & 52.6   & 56.6   & 52.6 & 56.5 \\
Naive unmixing  & 67.5 & 91.5 & 65.0 & 89.0 \\
\midrule
FinSNet~\cite{yoo2020finsnet}\textsuperscript{$\dagger$} & 84.5 & 75.5 & -- & -- \\
Overlap-aware inpainting + FT        & \textbf{92.0} & \textbf{95.5 }& \textbf{93.0} & \textbf{96.5 }\\
\bottomrule
\end{tabular}
{\footnotesize $\dagger$ FinSNet uses an older VeriFinger SDK.}
\vspace{-0.4cm}
\end{table}



\subsection{Minutiae Correspondence Analysis}
\begin{table*}[t]
\centering
\caption{Results of minutiae correspondence analysis. \textbf{Correspondence (\%)} measures the percentage of matched minutiae between probe and template. \textbf{Correspondence in overlap (\%)} and \textbf{Correspondence in non-overlap (\%)} report correspondence percentage restricted to minutiae in overlap and non-overlap regions, respectively. Best value per row per dataset shown in \textbf{bold}.}
\label{tab:minutiae_corr}
\small
\begin{tabular}{@{}l ccc ccc@{}}
\toprule
& \multicolumn{3}{c}{\textbf{TSOF}} & \multicolumn{3}{c}{\textbf{TLOF}} \\
\cmidrule(lr){2-4} 
\cmidrule(lr){5-7}
\textbf{Metric} & \makecell{Naive\\Unmixing} & \makecell{Overlap-aware\\Inpainting} & \makecell{+ Fine-tuning} & \makecell{Naive\\Unmixing} & \makecell{Overlap-aware\\Inpainting} & \makecell{+ Fine-tuning} \\
\midrule
Total minutiae in probes & 11,025 & 8,445	& 7,723 & 12,229 & 12,237 & 11,984\\
Correspondence (\%) & 32.00 & 46.34 & \textbf{48.94} & 36.05 & 36.55 &	\textbf{38.70}
 \\
Correspondence in overlap (\%) & 27.73 &	44.63 &	\textbf{47.54} & 21.68 &	31.59 &	\textbf{35.23}
 \\
Correspondence in non-overlap (\%) & 48.83 & \textbf{51.72} &	51.61 & \textbf{48.87} & 39.77 &	40.58 \\

\bottomrule
\end{tabular}
\vspace{-0.2cm}
\end{table*}

To investigate whether the proposed method faithfully reconstructs the ridge patterns, we analyze the percentage of minutiae correspondences (output by MegaMatcher) between the reconstructed probes and their corresponding mated templates in the gallery. These results are summarized in Table \ref{tab:minutiae_corr} for both TSOF and TLOF datasets. We also report the total number of minutiae extracted from the all probes. For further insights, we also breakdown the percentage of minutiae correspondences for both the overlapping and non-overlapping regions. In general, we expect the percentage of minutiae correspondences to increase (relative to naive unmixing) after separation of the overlapped fingerprint using the proposed method, because the reconstructed target fingerprints are expected to have a lower number of false/spurious minutiae (especially in the overlapping region). Furthermore, we expect this increase to be more pronounced in the overlapping regions. Both these hypotheses are clearly validated by the results in Table \ref{tab:minutiae_corr}. 

For the non-overlapping region, the percentage of minutiae correspondences should not change after reconstruction because ideally the non-overlapping region must be left unmodified by the separation algorithm. However, we observe that this percentage increases marginally in the TSOF dataset and decreases in the TLOF dataset. This could be due to the overall lower quality of images in the TLOF dataset, which makes it harder for the proposed model to faithfully reconstruct the ridges in the non-overlapping regions. This hypothesis is also confirmed by the fact that the total minutiae count decreases significantly for the TSOF dataset, a similar reduction is not observed for the TLOF dataset (possibly due to introduction of spurious minutiae in the non-overlapping region). This calls for better inpainting strategies that leave the non-overlapping regions unmodified even for low quality overlapped fingerprints. 

\subsection{Discussion and Limitations}

\noindent \textbf{Importance of progressive learning.} A natural question that has been left unaddressed earlier is the need for progressive learning. Indeed, it is possible to skip Stage 0 and/or Stage 1 in the proposed approach and directly update the base SD1.5 model using Stage 2 training so that the model directly learns to reconstruct component fingerprints from the overlapped fingerprint. Table~\ref{tab:ablation} shows that each stage contributes additional knowledge into the model, improving the reconstruction of the component fingerprints.

\begin{table}[h]
\centering
\caption{Closed-set identification performance based on different configurations of the proposed multi-stage training.}
\label{tab:ablation}
\small
\begin{tabular}{@{}l cc cc@{}}
\toprule
& \multicolumn{2}{c}{\textbf{TSOF}} & \multicolumn{2}{c}{\textbf{TLOF}} \\
\cmidrule(lr){2-3} \cmidrule(lr){4-5}
\textbf{Method} & TPIR & FPIR & TPIR & FPIR \\
\midrule
\multicolumn{5}{@{}l}{\textit{Overlap-aware inpainting}} \\
\quad Stage 2 only & 79.0 & 0.11 & 94.0 & 0.00 \\
\quad Stage 0 + 2 & 87.5 & 0.06 & 94.5 & 0.18 \\
\quad All stages & 89.0 & 0.17 & 93.5 & 0.05 \\
\midrule
\multicolumn{5}{@{}l}{\textit{+ fine-tuning}} \\
\quad Stage 2 only & 84.0 & 0.33 & 94.5 & 0.09 \\
\quad Stage 0 + 2  & 89.5 & 0.06 & 94.5 & 0.00 \\
\quad All stages & \textbf{90.5} & 0.17 & \textbf{95.0} & 0.00 \\

\bottomrule
\end{tabular}
\vspace{-0.1cm}
\end{table}

\begin{figure}[h]
    \centering
    \begin{subfigure}[b]{0.49\columnwidth}
        \includegraphics[width=\linewidth]{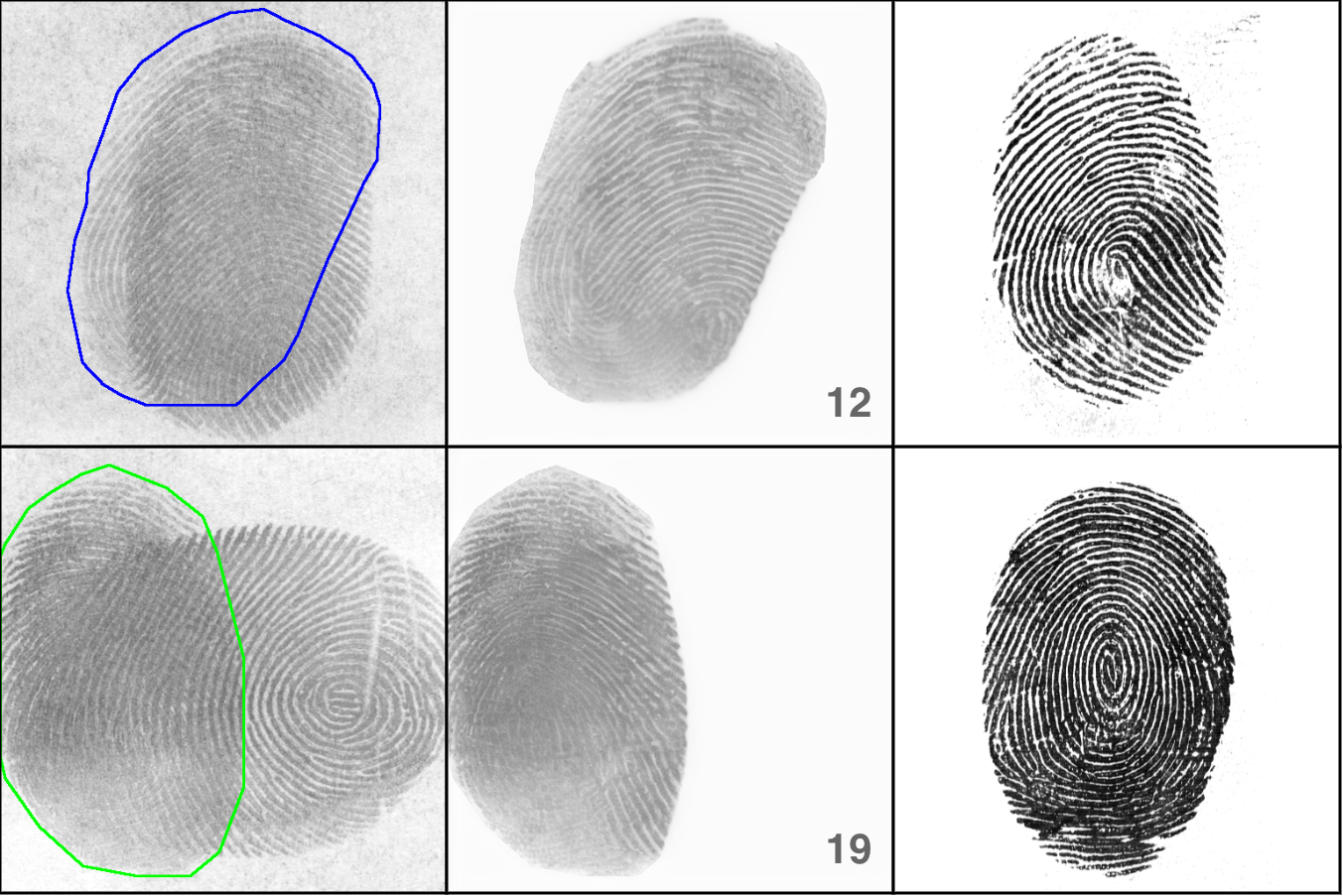}
        \caption{TLOF}
        \label{fig:fail_a}
    \end{subfigure}
    \begin{subfigure}[b]{0.49\columnwidth}
        \includegraphics[width=\linewidth]{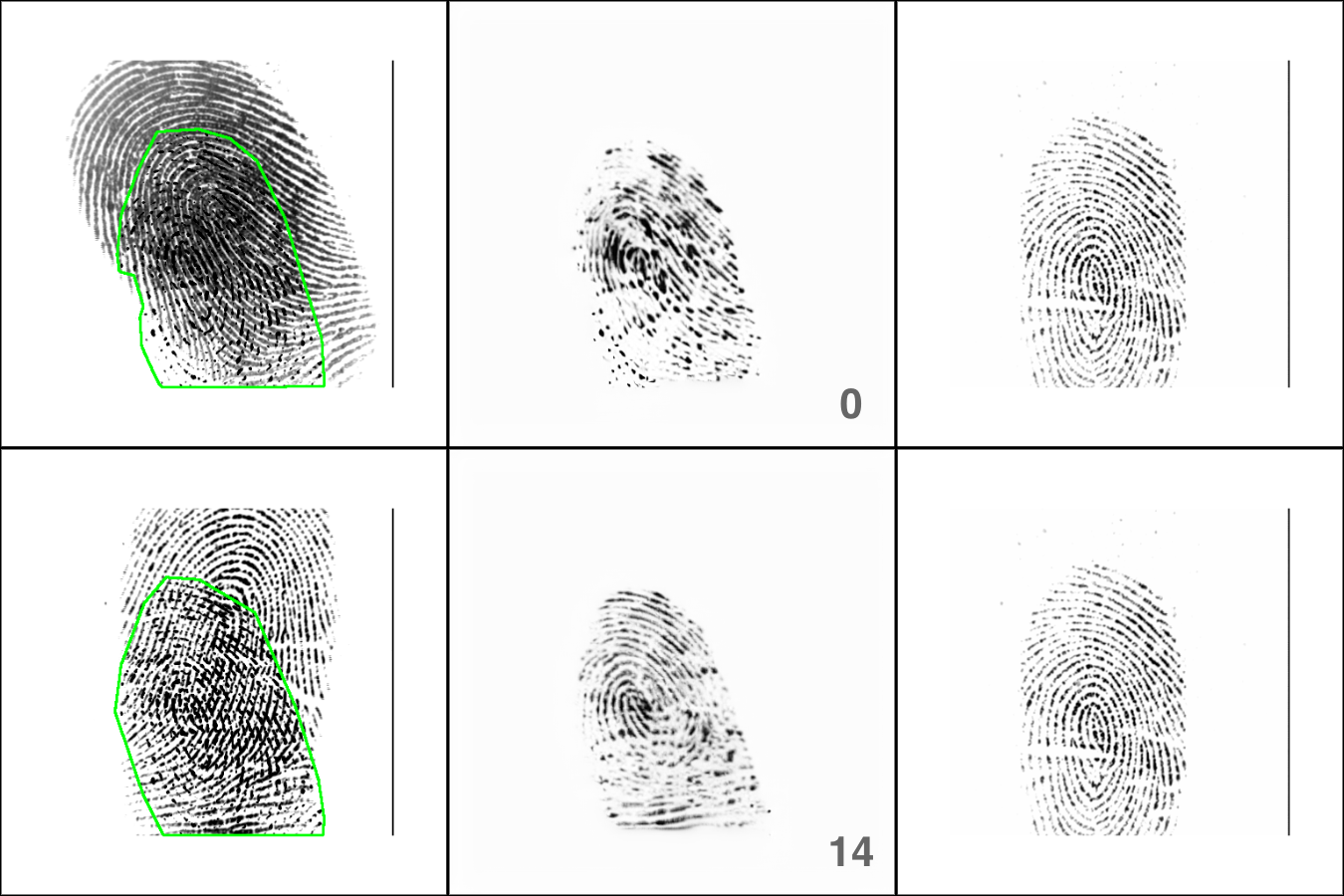}
        \caption{TSOF}
        \label{fig:fail_b}
    \end{subfigure}
    \caption{Examples of failure to reconstruct the target fingerprint accurately from (a) TLOF and (b) TSOF datasets. In each subfigure, the first column is the overlapped fingerprint with the target fingerprint outlined, second column is the reconstructed target (on the bottom right is the match score), and the last column is the mated template in the gallery.}
    \label{fig:failure_cases}
\end{figure}

\noindent \textbf{Limitations.} First, we investigate cases where the proposed method fails to reconstruct target fingerprints accurately. As shown in Figure \ref{fig:failure_cases}, when the ridge patterns of the target component print are of very low quality in the given overlapped fingerprint image due to the large overlap ratio and the strong domination of the other component print, it becomes difficult to reconstruct the target precisely. Thus, the proposed method needs further improvements to handle more difficult cases of overlap. Another limitation of the proposed method is the assumption that component masks $M_A$ and $M_B$ are available as additional input. Although this may be feasible in forensic case work, it hinders fully automated deployment. Developing an automated segmentation model to extract these masks is a natural future direction.   

\vspace{-0.2cm}
\section{Conclusion}

We formulated overlapped friction ridge separation as a conditional diffusion inpainting task and instantiated it with a stacked LoRA architecture that combines a frozen fingerprint generative prior, a single-fingerprint inpainting prior, and an overlap-aware multi-channel conditioning scheme. Trained on a large amount of synthetically overlapped fingerprints constructed from real plain fingerprint impressions, our model recovers the component fingerprints from the given overlapped print with reasonable precision. Minutiae correspondence analysis shows that our reconstructions roughly double the proportion of correctly recovered minutiae inside the overlapping region compared to naive separation based on masks. Future work includes detection and instance segmentation of overlapped fingerprint images.

\section{Acknowledgment}
This work was supported by the Center for Identification Technology Research (CITeR) Project \#25F-01M-SP.

{\small
\bibliographystyle{ieee}
\bibliography{egbib}
}

\clearpage
\newpage

\section{Supplementary Material}

\subsection{Qualitative Comparison With Baselines}

\begin{figure}[h]
    \centering
    \includegraphics[width=0.95\columnwidth]{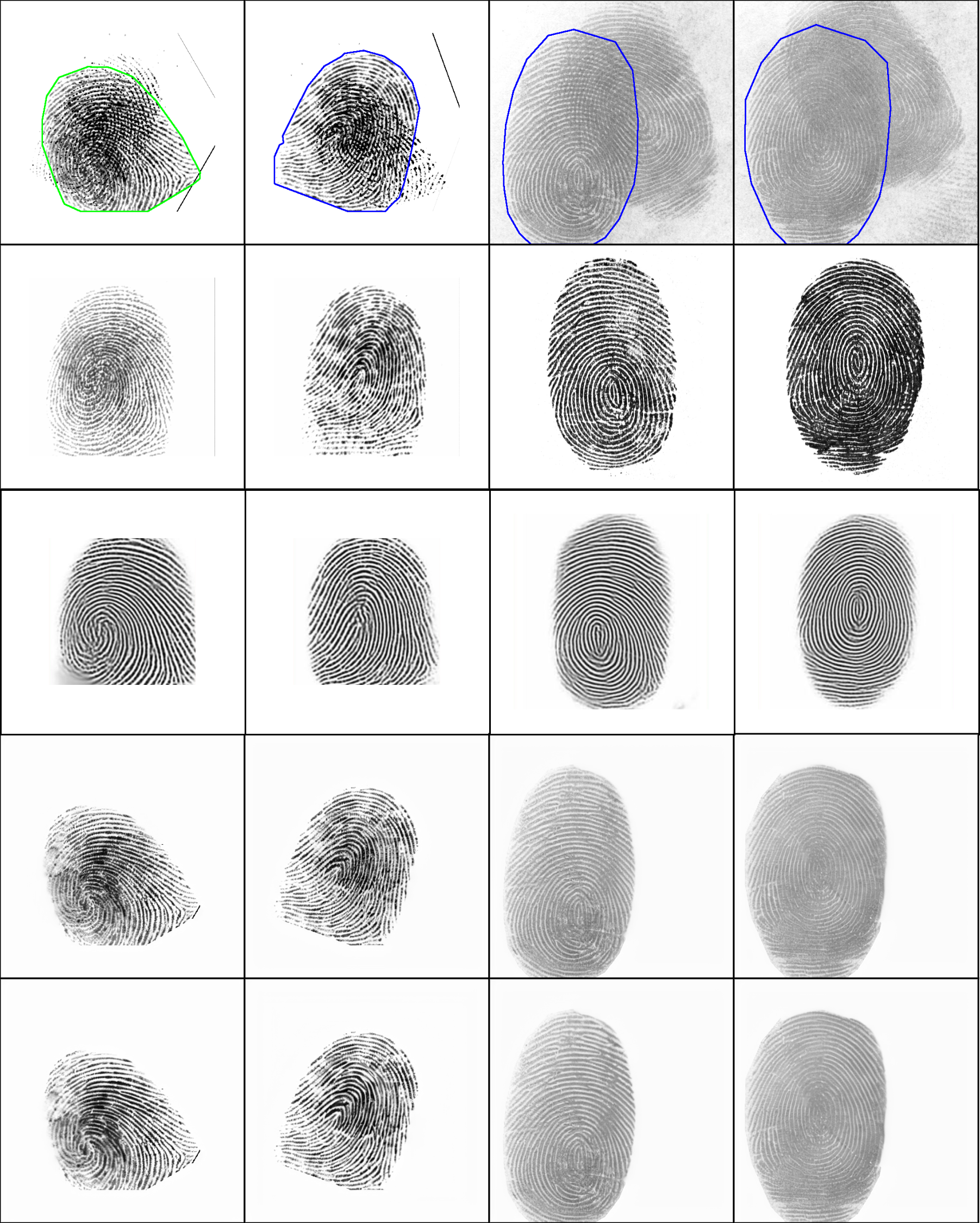}
    \caption{Qualitative comparison of the proposed method with previous learning-based reconstruction method. Rows from top to bottom are: overlapped fingerprint image with target fingerprint outlined, corresponding ground truth template in the gallery, reconstructed target fingerprint based on the FinSNet method~\cite{yoo2020finsnet}, reconstructed target fingerprint from our overlap-aware inpainting method without fine-tuning, and reconstructed target fingerprint from our overlap-aware inpainting method with fine-tuning.}
    \label{fig:comp_2_prev_work_finsnet}
\end{figure}

\begin{figure*}[h]
    \centering
    \includegraphics[width=0.76\linewidth]{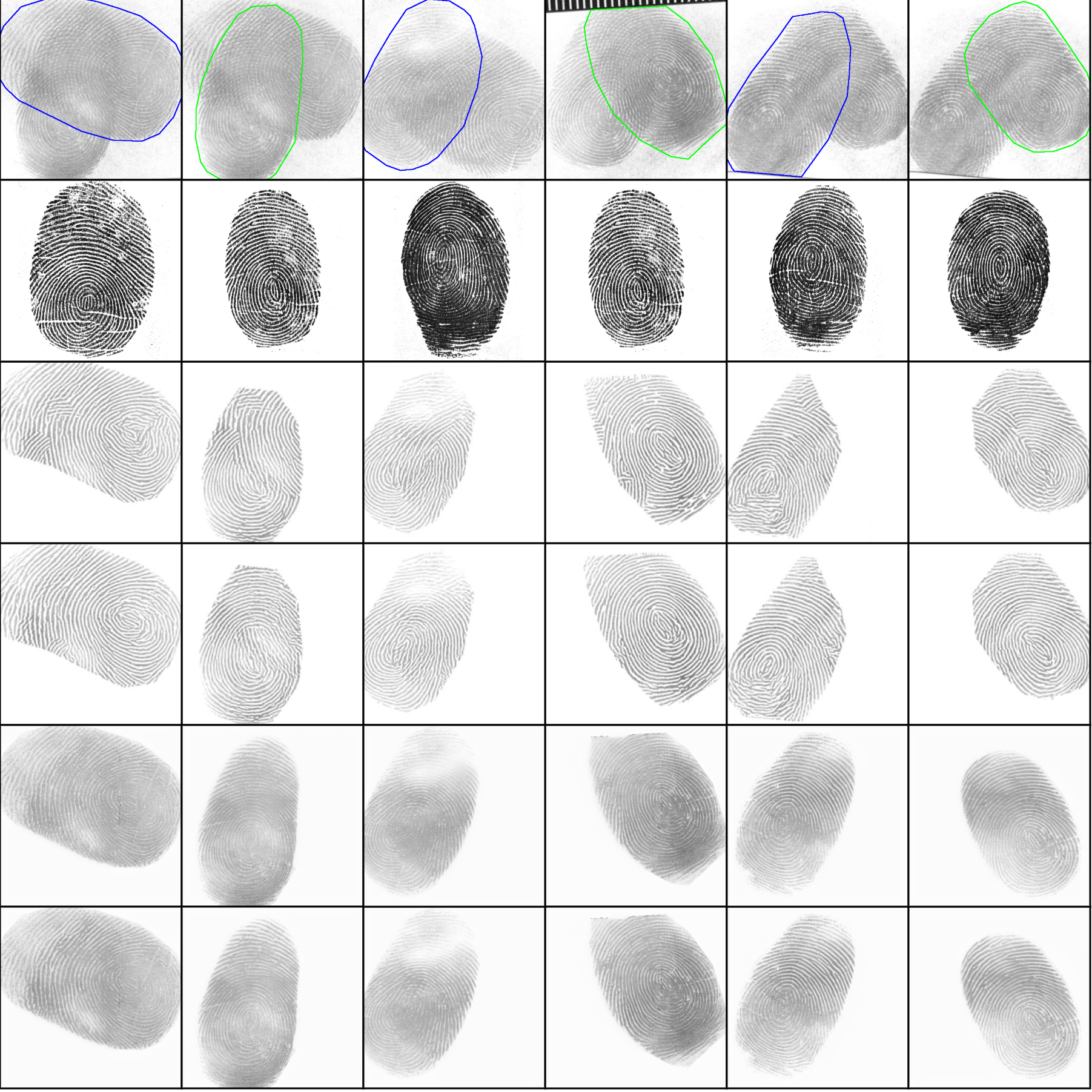}
    \caption{Qualitative comparison of the proposed method with previous model-based reconstruction methods. Rows from top to bottom are: overlapped fingerprint image with target fingerprint outlined, corresponding ground truth template in the gallery, reconstructed target fingerprint from~\cite{chenjain2011separating}, reconstructed target fingerprint from~\cite{feng2012robust}, reconstructed target fingerprint from our overlap-aware inpainting method without fine-tuning, and reconstructed target fingerprint from our overlap-aware inpainting method with fine-tuning.}
    \label{fig:comp_2_prev_work}
\end{figure*}

In Figure \ref{fig:comp_2_prev_work}, we compare the reconstructed fingerprint images obtained using the proposed method with the results of model-based reconstruction methods reported in~\cite{chenjain2011separating, feng2012robust}. While the first row shows the overlapped fingerprint image with the target fingerprint outlined, the second row shows the corresponding template in the gallery. Outputs of the methods in~\cite{chenjain2011separating} and~\cite{feng2012robust} are shown in rows 3 and 4, respectively. Finally, the results of our overlap-aware inpainting method without and with fine-tuning are shown in rows 5 and 6, respectively. It can be observed that our method produces component fingerprints that are more realistic and have the correct ridge structure in comparison to the template images shown in row 2. In Figure \ref{fig:comp_2_prev_work_finsnet}, we further qualitatively compare our results with the FinSNet~\cite{yoo2020finsnet} method on both TSOF and TLOF. Once again, it can be observed that the proposed approach more faithfully reconstructs the target fingerprint (in both ridge structure and style) compared to the FinSNet~\cite{yoo2020finsnet} method.

\begin{figure}[h]
    \centering
    \includegraphics[width=0.9\columnwidth]{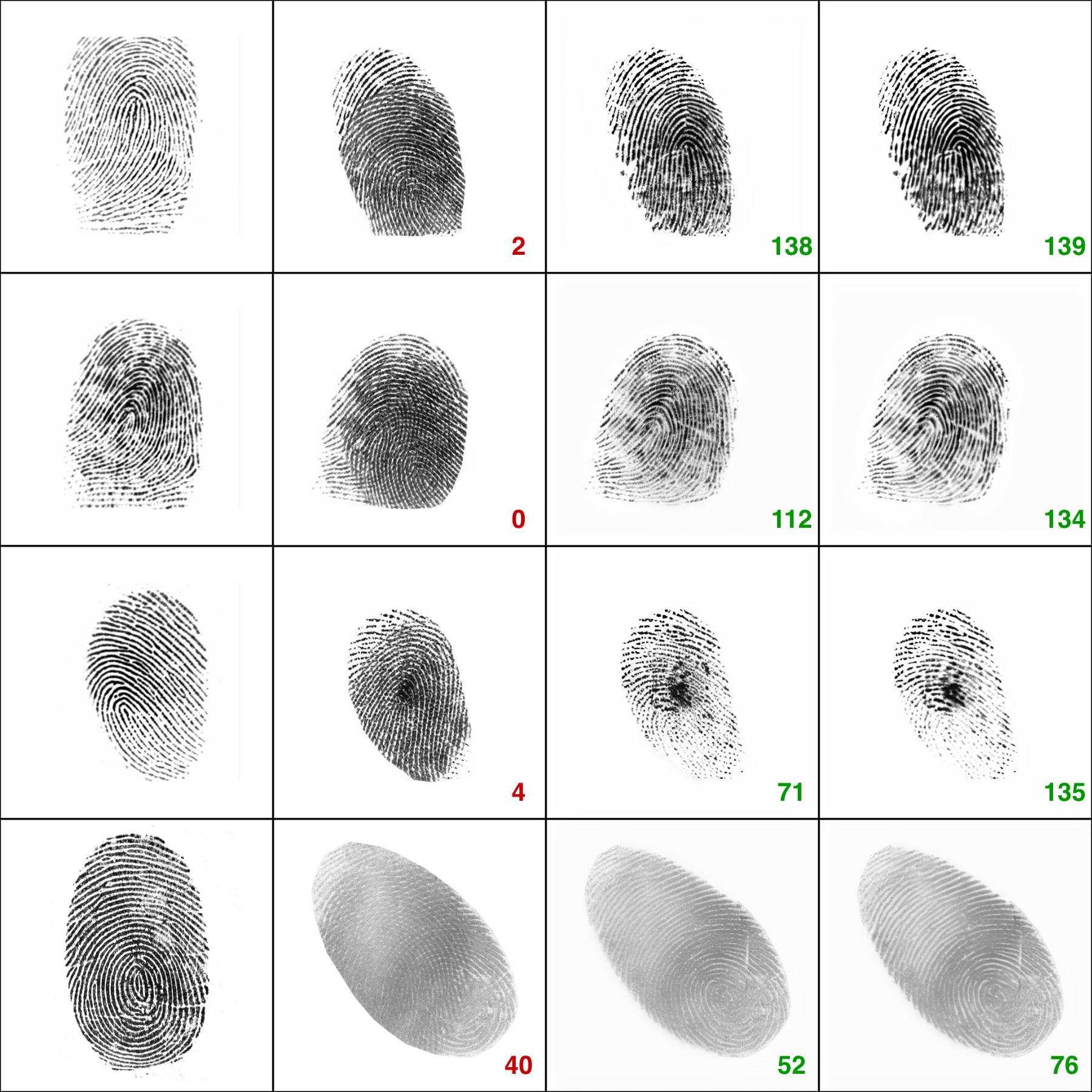}
    \caption{Examples of cases where the proposed method achieves successful match in the verification setting. The match scores are indicated on the bottom right of every image. Columns left to right are: mated template in the gallery, reconstructed target fingerprint based on naive unmixing, reconstructed target fingerprint from our overlap-aware inpainting method without fine-tuning, and reconstructed target fingerprint from our overlap-aware inpainting method with fine-tuning.}
    \label{fig:reject_2_accept}
\end{figure}

\subsection{Verification Results}

\begin{figure}[h]
    \centering
    \includegraphics[width=0.95\columnwidth]{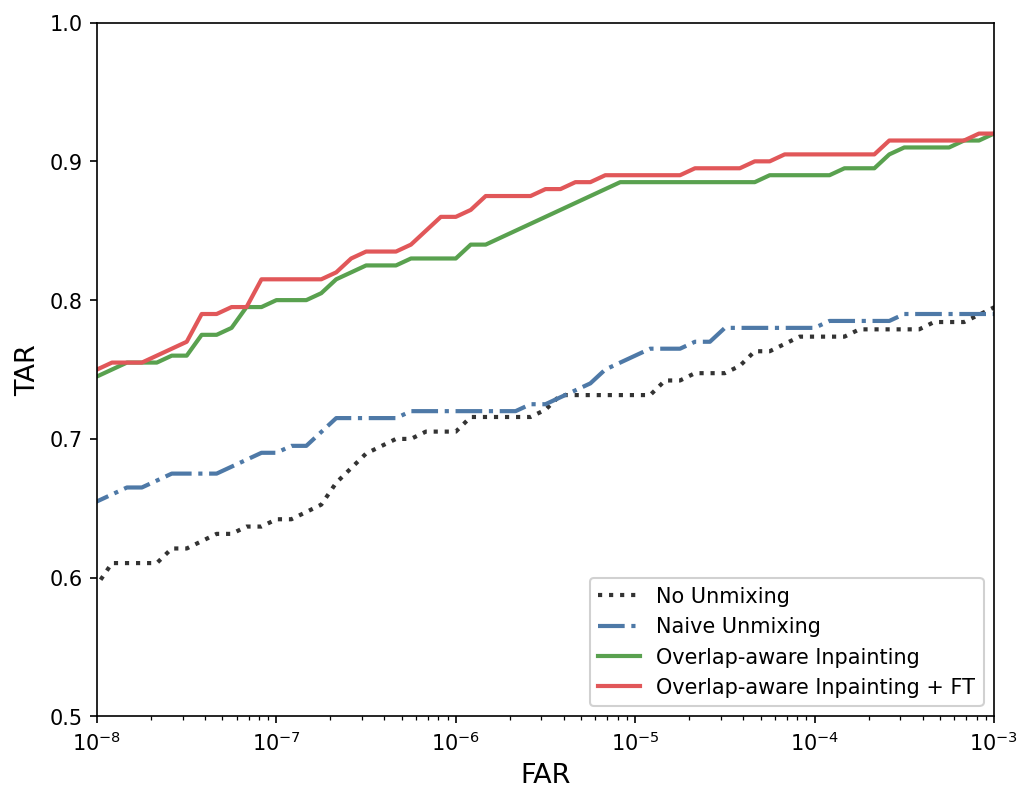}
    \caption{ROC curves for different benchmarking scenarios on the TSOF dataset.}
    \label{fig:roc}
\end{figure}

In Figure \ref{fig:roc}, we report receiver operating characteristic (ROC) curves for 4 benchmarking scenarios on the TSOF dataset. Note that no impostor matches have been conducted for this plot. The thresholds for different FMR values are obtained from the MegaMatcher SDK manual and the TMR values at these thresholds are computed. We can observe from the plot that the proposed overlap-aware inpainting method significantly improves the verification accuracy of separated fingerprints and the fine-tuning step provides some further marginal benefits. Examples of cases where the proposed method led to successful match are shown in Figure \ref{fig:reject_2_accept}. These examples clearly show that the proposed method can substantially increase the match score of the reconstructed probe against the corresponding mated template in the gallery. 

\begin{table}[h]
\centering
\caption{Sensitivity of verification performance to overlap ratio. The values correspond to TMR (\%) @ FMR=0.01\%.} 
\label{tab:overlap_ratio}
\small
\setlength{\tabcolsep}{4pt}
\renewcommand{\arraystretch}{1.05}
\begin{tabular}{@{}c ccc@{}}
\toprule
\textbf{Overlap \%} & \makecell{\textbf{Naive} \\ \textbf{Unmixing}} & \makecell{\textbf{Overlap-aware}\\\textbf{Inpainting}} & \textbf{+ Fine-tuning} \\
\midrule
15 & 100.0 & 100.0 & 100.0 \\
30 & 100.0 & 100.0 & 100.0 \\
45 & \phantom{0}99.5 & \phantom{0}99.0 & \phantom{0}98.5 \\
50 & \phantom{0}99.5 & \phantom{0}99.0 & \phantom{0}99.0 \\
65 & \phantom{0}90.5 & \phantom{0}93.5 & \phantom{0}93.5 \\
80 & \phantom{0}75.5 & \phantom{0}89.0 & \phantom{0}89.5 \\
\bottomrule
\end{tabular}
\vspace{-0.1cm}
\end{table}
\subsection{Sensitivity to overlap ratio} 

To characterize how separation performance varies with problem difficulty (measured in terms of the overlap ratio), we construct a controlled evaluation set from FVC2002-DB1 identities 101--110 that mimics the TSOF dataset but with six overlap ratio buckets (15--80\%). As shown in Table \ref{tab:overlap_ratio}, MegaMatcher is sufficiently robust to handle cases where the overlap ratio is less than $30$\% leading to $100$\% TMR at $0.01$\% FMR. Note that in the identification setting, there may still be some false positives. The drop in TMR is not precipitous even up to an overlap ratio of $50$\%. When the overlap ratio exceeds $50$\%, the performance degradation becomes significant and the benefits of the proposed method become more pronounced.

\end{document}